\documentclass{article}

\usepackage{arxiv}

\usepackage[utf8]{inputenc} % allow utf-8 input
\usepackage[T1]{fontenc}    % use 8-bit T1 fonts
\usepackage{hyperref}       % hyperlinks
\usepackage{url}            % simple URL typesetting
\usepackage{booktabs}       % professional-quality tables
\usepackage{amsfonts}       % blackboard math symbols
\usepackage{nicefrac}       % compact symbols for 1/2, etc.
\usepackage{microtype}      % microtypography
\usepackage{lipsum}
\usepackage{graphicx}
\graphicspath{ {./images/} }
\usepackage{amsmath}
\usepackage[ruled,vlined]{algorithm2e}
 
\title{Handcrafted Feature--Assisted One-Class Learning for Artist Authentication in Historical Drawings}

\author{
  Hassan Ugail\\
  Centre for Visual Computing and Intelligent Systems \\
  University of Bradford \\
  United Kingdom\\
  \texttt{h.ugail@bradford.ac.uk} \\
  \And
  Jan Ritch-Frel\\
  Independent Media Institute \\
  United States\\
  \And
  Irina Matuzava\\
  Independent Media Institute \\
  United States\\
  \\
}

\begin{document}
\maketitle

\begin{abstract}
Authentication and attribution of works on paper remain persistent challenges in cultural heritage, particularly when the available reference corpus is small and stylistic cues are primarily expressed through line and limited tonal variation. We present a verification-based computational framework for historical drawing authentication using one-class autoencoders trained on a compact set of interpretable handcrafted features. Ten artist-specific verifiers are trained using authenticated sketches from the Metropolitan Museum of Art open-access collection, the Ashmolean Collections Catalogue, the Morgan Library and Museum, the Royal Collection Trust (UK), the Victoria and Albert Museum Collections, and an online catalogue of the Casa Buonarroti collection and evaluated under a biometric-style protocol with genuine and impostor trials. Feature vectors comprise Fourier-domain energy, Shannon entropy, global contrast, GLCM-based homogeneity, and a box-counting estimate of fractal complexity. Across 900 verification decisions (90 genuine and 810 impostor trials), the pooled system achieves a True Acceptance Rate of 83.3\% with a False Acceptance Rate of 9.5\% at the chosen operating point. Performance varies substantially by artist, with near-zero false acceptance for some verifiers and elevated confusability for others. A pairwise attribution of false accepts indicates structured error pathways consistent with stylistic proximity and shared drawing conventions, whilst also motivating tighter control of digitisation artefacts and threshold calibration. The proposed methodology is designed to complement, rather than replace, connoisseurship by providing reproducible, quantitative evidence suitable for data-scarce settings common in historical sketch attribution.
\end{abstract}


\section{Introduction}

\subsection{Context and motivation}

The authentication and attribution of historical artworks are central concerns in art history, conservation, and the art market. For works on paper, these concerns are intensified by the material and documentary conditions under which drawings survive: sketchbooks may be dispersed, sheets may be trimmed or mounted, and many works exist in multiple states or workshop contexts. Connoisseurship remains indispensable in this domain, yet it is intrinsically difficult to formalise, reproduce, and quantify, particularly when disputes arise, and decision-makers require transparent evidence beyond expert opinion \cite{johnson2008image,stork2020computer}.

Computational methods offer a complementary mode of evidence. In particular, the evaluation logic of biometric verification provides an appealing analogue: an input sample is verified against a target identity, and system performance is characterised by false acceptance and false rejection under explicit operating points. In art authentication, the ``identity'' is the target artist, and impostor trials represent non-target artists (and, in principle, forgeries). This framing aligns with the open-set nature of attribution, where it is rarely possible to enumerate all plausible non-target classes.

Recent advances in machine learning have demonstrated impressive results for art attribution when large labelled datasets are available. A comprehensive review of the use of artificial intelligence in art authentication is provided by Cetinic and She (2022), who document the field’s shift from traditional computer vision methods toward deep neural networks \cite{cetinic2022understanding}. Building on this trend, vision transformers have been applied to artist attribution, with self-attention mechanisms enabling state-of-the-art performance \cite{madhu2023recognizing}. In parallel, graph neural networks have been shown to effectively model stylistic relationships between artworks, capturing higher-level structural dependencies beyond pixel-based features \cite{castellano2021deep}. Despite these advances, such approaches generally depend on thousands of examples per artist, a requirement that is rarely met in the context of historical sketches.

\subsection{Data scarcity and the rationale for one-class verification}

Deep supervised attribution methods generally require large labelled datasets and benefit from broad negative sampling \cite{ugail2023raphael}. Historical sketches rarely satisfy these conditions. Even for major artists, the number of authenticated drawings available as consistent digital surrogates is limited; moreover, intra-artist variability can be substantial because sketches are often rapid, exploratory studies rather than finished works \cite{elgammal2018shape}. These constraints motivate one-class learning, where the model learns a representation of the authentic distribution of a single artist and flags deviations as anomalous \cite{ruff2021unifying, pang2021deep}. One-class verification is particularly appropriate when negative classes are heterogeneous, incompletely characterised, or strategically adversarial (as in forgery scenarios) \cite{geng2020recent}.

The challenge of limited training data is not unique to art authentication. Few-shot learning approaches have been explored in various domains \cite{strezoski2021learning}, but these typically still require more examples than are available for many historical artists. Transfer learning from pre-trained models offers another avenue \cite{sabatelli2019deep}, but the domain gap between natural images and historical sketches can be substantial. One-class learning sidesteps these issues by focusing solely on modelling the authentic distribution without requiring comprehensive negative examples.

\subsection{The role of handcrafted features in data-scarce settings}

Whilst end-to-end deep learning has dominated recent work in computer vision, handcrafted features retain important advantages in data-scarce scenarios. They reduce sample complexity through dimensionality reduction, provide interpretability enabling expert validation, incorporate domain knowledge, and offer robustness to distribution shift \cite{impett2017totentanz}. For sketch authentication, where colour information is limited and style is expressed primarily through line work, tonal distribution, and mark-making patterns, carefully designed handcrafted features can capture essential stylistic signals whilst remaining trainable with minimal data.

Texture features derived from Grey-Level Co-occurrence Matrices have proven effective for distinguishing artistic techniques \cite{haralick1973textural, bianco2020predicting}. Frequency domain analysis reveals characteristic patterns in how artists distribute spatial frequencies \cite{pouli2020color, lyu2019identifying}. Fractal analysis captures the hierarchical complexity of mark-making \cite{mandelbrot1982fractal, taylor2019perceptual, sigaki2020history}. Information-theoretic measures quantify tonal complexity and distributional properties \cite{shannon1948mathematical, rigau2021conceptualizing}. By combining complementary features operating at different scales, we can construct compact yet informative representations suitable for one-class learning.

\subsection{Scope and contributions}

This study develops a reproducible verification framework for sketch authentication under severe data constraints. The central methodological contribution is an artist-specific, one-class autoencoder verifier trained on interpretable handcrafted features that are well suited to line-dominant media. The empirical contribution is a multi-artist evaluation across ten historical artists, reporting both pooled and artist-specific biometric metrics with Wilson confidence intervals, along with a structured attribution of false-accept pathways to identify systematic confusability between artists.

Our principal contributions are as follows:
\begin{itemize}
\item A novel application of one-class autoencoder architecture to historical sketch authentication, demonstrating effective discrimination despite severe data scarcity (20 training images per artist).
\item Identification and formal definition of five complementary handcrafted features—Fourier energy, Shannon entropy, contrast, GLCM homogeneity, and box-counting fractal dimension—that effectively capture artistic style in sketch media.
\item Comprehensive multi-artist evaluation using a rigorous biometric verification framework with 900 trials, reporting performance metrics with Wilson binomial confidence intervals appropriate for small sample sizes.
\item Pairwise confusion attribution revealing structured error pathways that provide insights into artistic relationships and stylistic proximity.
\item A methodological framework suitable for extension to other historical artists and potentially applicable to other domains of art authentication where training data is inherently limited.
\end{itemize}

\section{Related Work}

\subsection{Machine learning in art authentication}

Research on computational art analysis spans traditional feature engineering, modern deep learning, and hybrid approaches. Surveys in computer vision applied to art have highlighted both the promise of machine learning for attribution and the practical barriers posed by limited data, domain shift, and the interpretability gap between learned features and art-historical concepts \cite{johnson2008image, stork2020computer}. Early work by \cite{li2004classification} applied wavelet analysis to paintings, demonstrating that computational methods could detect stylistic patterns not readily apparent to human observers.

Recent advances in deep learning have shown impressive results when sufficient data is available. \cite{castellano2021deep} provided a comprehensive overview of deep learning approaches to pattern extraction in paintings and drawings, achieving accuracies exceeding 90\% on datasets containing thousands of images per artist. \cite{madhu2023recognizing} applied vision transformers to character recognition in art history, leveraging self-attention mechanisms to capture long-range dependencies. \cite{cetinic2022understanding} surveyed the broader landscape of AI in art, identifying three major approaches: supervised classification using convolutional neural networks, transfer learning from pre-trained models, and generative modelling for anomaly detection.

However, these impressive results typically require substantial training data. \cite{saleh2015large} curated large-scale art datasets containing thousands of images per artist category, enabling training of deep networks. \cite{strezoski2021learning} explicitly addressed the data efficiency challenge, proposing few-shot learning approaches that could learn from as few as five examples per class through meta-learning, though performance remained below that achieved with larger datasets. \cite{sabatelli2019deep} investigated transfer learning for art classification, demonstrating that pre-trained features could partially address data scarcity, but noted substantial performance gaps compared to in-domain training.

For specific authentication challenges, targeted approaches have proven effective. \cite{garcia2020contextnet} developed methods for painting classification and retrieval that account for contextual information. \cite{jiang2020learning} applied deep networks with data augmentation to forgery detection. \cite{lyu2019identifying} used computer-generated image detection techniques for art authentication. Most existing work has focused on oil paintings, with limited attention to works on paper, motivating the current study's focus on sketch authentication.

\subsection{Autoencoders and anomaly detection}

In parallel, the anomaly detection literature has formalised one-class learning and reconstruction-based scoring, with autoencoders remaining a standard approach when the objective is to model ``normal'' data and detect deviations \cite{ruff2021unifying, pang2021deep}. Autoencoders, first introduced by \cite{rumelhart1986learning} and refined over decades, learn to compress input data into lower-dimensional latent representations and then reconstruct the original input. By training exclusively on normal data, the autoencoder becomes specialised at reconstructing similar instances; anomalous data produces higher reconstruction errors, providing a quantitative anomaly score.

\cite{ruff2021unifying} provided a comprehensive survey of deep learning approaches to anomaly detection, identifying autoencoders as particularly suitable for one-class learning scenarios where normal data is abundant but anomalous data is scarce or unknown. Their taxonomy distinguished between reconstruction-based methods, which use reconstruction error as an anomaly score, and embedding-based methods, which learn compact representations optimised for separating normal from anomalous data. \cite{pang2021deep} conducted an extensive review highlighting advances in convolutional autoencoders, variational autoencoders, and adversarial training approaches, noting that reconstruction-based methods remain effective for high-dimensional data such as images.

Recent work has addressed training stability and performance optimisation. \cite{gong2019memorizing} proposed memory-augmented autoencoders that explicitly store prototypical normal patterns. \cite{liu2020towards} developed constrained autoencoders that regularise the latent space to ensure meaningful representations. \cite{zhou2021attention} introduced attention-based autoencoders that selectively focus on salient regions, improving discrimination between normal and anomalous patterns.

In medical imaging, a domain with parallels to art authentication regarding limited pathological examples and high-dimensional data, autoencoders have proven highly effective. \cite{baur2021deep} applied deep autoencoders to brain MRI analysis, achieving state-of-the-art anomaly detection for identifying pathological changes. \cite{schlegl2019fanoGAN} developed generative adversarial network-based approaches for retinal imaging. \cite{alshehri2023breast} combined texture features with deep learning for breast cancer classification, achieving 98.6\% accuracy by leveraging handcrafted features alongside learnt representations—a hybrid approach that informed our methodology.

\subsection{Feature engineering for art analysis}

For works on paper, feature selection is consequential. Sketches provide less colour information than paintings and often express style through mark-making, tonal distribution, and compositional density. Consequently, interpretable features derived from frequency analysis, information theory, and texture statistics can be advantageous in data-scarce settings because they reduce dimensionality whilst retaining meaningful stylistic signal.

Texture features derived from Grey-Level Co-occurrence Matrices (GLCM) remain widely used. \cite{haralick1973textural} introduced these features for image classification, defining measures including contrast, homogeneity, energy, and entropy that capture statistical properties of spatial relationships between pixel intensities. \cite{bianco2020predicting} demonstrated that GLCM features combined with modern classifiers could achieve competitive performance for texture classification, including artistic textures. They identified homogeneity and entropy as particularly discriminative for distinguishing between artistic techniques. \cite{cusano2020combining} extended texture analysis to paintings, showing that multi-scale texture features could capture brushstroke characteristics across different levels of detail.

Frequency domain analysis continues to provide insights into artistic style. \cite{lyu2019identifying} applied wavelet analysis to detect art forgeries, demonstrating that authentic works exhibit characteristic frequency distributions that forgeries struggle to replicate. \cite{pouli2020color} used Fourier analysis to characterise colour and luminance patterns in paintings, showing that different artistic movements exhibit distinctive frequency signatures. These findings supported our inclusion of Fourier energy as a discriminative feature for capturing the scale and regularity of mark-making.

Fractal analysis has evolved beyond simple dimension estimation to become a sophisticated tool for characterising artistic complexity. \cite{mandelbrot1982fractal} established the theoretical foundations of fractal geometry. \cite{taylor2019perceptual} refined fractal analysis for Jackson Pollock authentication, addressing criticisms of earlier work and demonstrating robust discrimination between authentic drip paintings and imitations. \cite{sigaki2020history} applied multifractal analysis to paintings across art history, revealing that fractal complexity evolved systematically over centuries and differed characteristically between artistic movements, validating fractal features as meaningful stylistic markers.

Information-theoretic features have gained renewed attention for art analysis. \cite{shannon1948mathematical} established the theoretical framework for information theory. \cite{rigau2021conceptualizing} applied entropy-based measures to quantify visual complexity in artworks, demonstrating that entropy captures perceptually meaningful aspects of artistic composition. \cite{marin2022examining} investigated aesthetic preferences using information-theoretic features, finding that moderate complexity and entropy correlate with aesthetic appeal.

Recent work has explored feature fusion approaches. \cite{montagner2021feature} combined multiple feature types for artwork analysis, demonstrating that complementary features capture different stylistic aspects. Their work influenced our selection of five distinct features operating at different scales and capturing different properties of artistic style. This paper therefore adopts a compact feature vector designed to capture complementary aspects of drawing structure and complexity whilst remaining interpretable to domain experts.

\subsection{Biometric verification frameworks}

Art authentication shares methodological parallels with biometric verification, where individuals are authenticated based on intrinsic characteristics. The biometric literature provides rigorous frameworks for evaluating authentication systems that translate naturally to art verification contexts. \cite{jain201650} provided a comprehensive introduction to biometric systems, defining standard metrics including False Acceptance Rate, False Rejection Rate, and Equal Error Rate. Their framework for evaluating one-to-many identification scenarios directly applies to art authentication where a work is compared against a database of known artists.

\cite{grother2020face} discussed evaluation methodologies for biometric systems under realistic conditions, emphasising the importance of appropriate confidence intervals when sample sizes are limited. Their advocacy for Wilson binomial intervals \cite{wilson1927probable, agresti1998approximate} over normal approximations influenced our statistical approach. \cite{beveridge2021report} addressed challenges in comparing biometric systems across different datasets and evaluation protocols, highlighting the need for standardised reporting of performance metrics.

The concept of impostor trials in biometric verification directly parallels the challenge of distinguishing an artist's genuine works from those by other artists. \cite{marcel2019handbook} provided an overview of presentation attack detection, addressing the problem of deliberate spoofing—analogous to forgery in art authentication. Their discussion of anomaly-based detection methods informed our one-class learning approach. The biometric framework's explicit treatment of operating points, trade-offs between false acceptance and false rejection, and evaluation under realistic trial structures provides a rigorous foundation for art authentication research.

\section{Materials and Methods}

\subsection{Dataset and curation}
We curated a dataset of $K=10$ artists. For each artist, $n_{\mathrm{train}}=20$ authenticated works were used for model training and $n_{\mathrm{test}}=9$ authenticated works were reserved for evaluation, yielding $|S_a|=29$ images per artist and $290$ images in total. The $20/9$ split reflects the practical constraints of data-scarce authentication and ensures that all reported results correspond to a single, fixed protocol.

Images were sourced from publicly available, open-access collections, including the Metropolitan Museum of Art’s online collection, the Ashmolean Collections Catalogue, the Morgan Library and Museum, the Royal Collection Trust (UK), the Victoria and Albert Museum Collections, and the Casa Buonarroti online catalogue. Selection of the images was restricted to drawings and sketches attributed to ten artists:
Anthonis van den Wijngaerde (c. 1510–1561, Flemish topographical artist), sourced from the Metropolitan Museum of Art and the Ashmolean Collections Catalogue; John Constable (1776–1837, English landscape painter), sourced from the Metropolitan Museum of Art and the Victoria and Albert Museum Collections; Giovanni Francesco Barbieri, also known as Guercino (1591–1666, Italian Baroque painter), sourced from the Metropolitan Museum of Art and the Ashmolean Collections Catalogue; John William Waterhouse (1849–1917, English Pre-Raphaelite painter), sourced from the Metropolitan Museum of Art and the Victoria and Albert Museum Collections; Michelangelo Buonarroti (1475–1564, Italian Renaissance master), sourced from the Metropolitan Museum of Art, the Morgan Library and Museum, the Royal Collection Trust (UK), and an online catalogue of the Casa Buonarroti collection; Raffaello Sanzio, known as Raphael (1483–1520, Italian Renaissance master), sourced from the Metropolitan Museum of Art and the Ashmolean Collections Catalogue; Thomas Sully (1783–1872, American portrait painter), sourced from the Metropolitan Museum of Art; William Trost Richards (1833–1905, American landscape painter), sourced from the Metropolitan Museum of Art; James McNeill Whistler (1834–1903, American tonalist painter), sourced from the Metropolitan Museum of Art; and Wilhelm Stettler (1643–1708, Swiss draughtsman), sourced from the Metropolitan Museum of Art and the Ashmolean Collections Catalogue.

The selection criteria aimed to minimise confounds: images were chosen to avoid palimpsests or multi-work sheets where possible, and were cropped to reduce borders and extraneous page context, thereby limiting the influence of mount tone, margins, and institutional photographing conventions. Works were selected from fully provenanced collections where attribution is supported by unchallenged scholarship and, where available, documentary evidence. The temporal and stylistic diversity—spanning Renaissance through nineteenth century, and including Italian, Flemish, English, American, and Swiss traditions—ensures that the comparison set presents genuine challenges for authentication.

For each artist, thirty images were selected. Twenty images were used for training (model development and parameter estimation), and ten images were held out for testing (performance evaluation). This 67:33 split provides sufficient training data for the autoencoder whilst reserving adequate test data for robust performance evaluation, given the constraints of data availability. The evaluation protocol used 90 trials per artist-specific model: 9 genuine trials (test images of the target artist) and 81 impostor trials (9 test images from each of the nine non-target artists), yielding 900 pooled decisions across ten models. The reported results therefore correspond to a test set of nine images per artist within the verification experiments, with one image per artist unavailable or excluded during final evaluation. This trial structure, including pooled counts and per-artist confusion matrices, is consistent across the evaluation artefacts used in this study.

\begin{figure}
\centering
\includegraphics[width=0.9\textwidth]{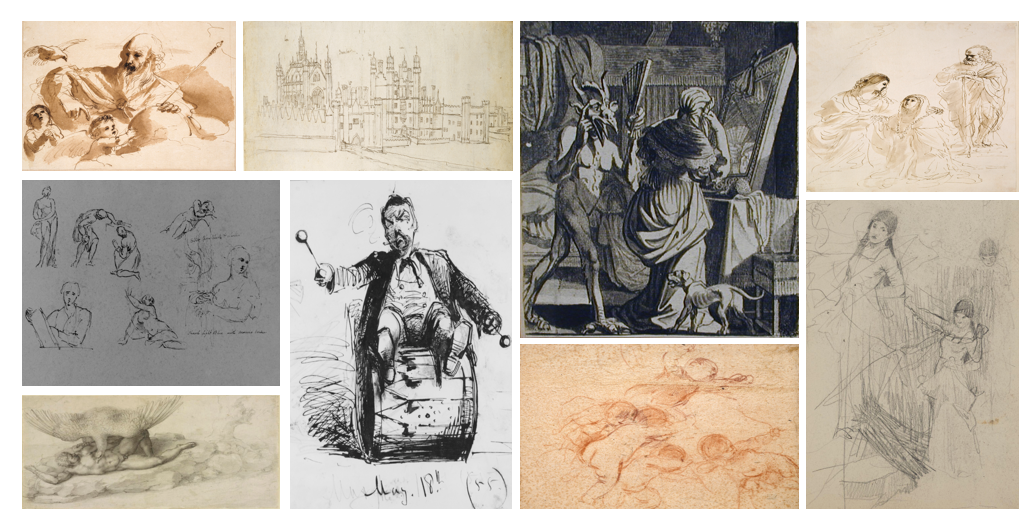}
\caption{Sample sketches from the dataset showing stylistic diversity across the ten artists.}
\label{fig:samples}
\end{figure}

\subsection{Preprocessing}

Each image is resized to $224 \times 224$ pixels using bicubic interpolation, which provides smooth interpolation appropriate for natural images and artwork. For features defined on luminance or texture statistics, images are converted to greyscale using the standard luminance transform
\begin{equation}
Y = 0.299R + 0.587G + 0.114B,
\end{equation}
which approximates human luminance perception by weighting the green channel most heavily. Intensities are normalised to $[0,1]$ by dividing by 255, ensuring consistent numeric ranges across features. This standardised preprocessing pipeline ensures that features are computed consistently across all images regardless of original resolution, colour depth, or digitisation protocols.

\subsection{Feature extraction}

For each image $i$, we compute a five-dimensional feature vector
\[
\mathbf{f}_i = [E_{\text{Fourier},i}, H_{\text{Shannon},i}, \sigma_{\text{contrast},i}, H_{\text{homogeneity},i}, D_{\text{BC},i}]^\top \in \mathbb{R}^5,
\]
where $D_{\text{BC}}$ denotes a box-counting estimate of fractal complexity (the Hausdorff-Besicovitch dimension). The features are chosen to capture complementary aspects of sketches: global frequency energy reflecting mark-making scale, tonal information content measuring distributional complexity, intensity dispersion quantifying value range, local spatial regularity characterising texture smoothness, and multi-scale edge complexity encoding hierarchical structure. These features operate at different scales and capture different properties of artistic style, providing a compact yet informative representation.

To reduce sensitivity to digitisation variability, all images are cropped to the artwork region during curation to remove page margins and mounts. No additional background masking is applied beyond this crop. Images are resized to $224\times224$ and normalised to $[0,1]$ prior to feature computation.

For GLCM features, greyscale intensities are uniformly quantised to $L_q=64$ levels prior to GLCM computation. Homogeneity is computed using distance $d=1$ pixel over orientations $\theta\in\{0^\circ,45^\circ,90^\circ,135^\circ\}$ and averaged to provide approximate rotation invariance.

For the box-counting fractal dimension estimate, edges are extracted using Canny edge detection with Gaussian smoothing $\sigma=1.0$ and hysteresis thresholds $t_{\mathrm{low}}=0.10$ and $t_{\mathrm{high}}=0.20$ (defined on the $[0,1]$ intensity scale). Box counts are computed over box sizes $\varepsilon \in \{2,4,8,16,32,64\}$ pixels, and the fractal dimension is estimated as the slope of a least-squares regression of $\log N(\varepsilon)$ on $\log(1/\varepsilon)$.

\subsubsection{Fourier energy}

Let $P(i,j)$ denote the greyscale image of size $M \times N$. The two-dimensional Discrete Fourier Transform (DFT) decomposes the image into frequency components:
\begin{equation}
F(u,v) = \sum_{i=0}^{M-1}\sum_{j=0}^{N-1} P(i,j)\exp\left[-2\pi\mathbf{i}\left(\frac{ui}{M}+\frac{vj}{N}\right)\right],
\end{equation}
where $u \in \{0,1,\ldots,M-1\}$ and $v \in \{0,1,\ldots,N-1\}$ are frequency indices, and $\mathbf{i}$ represents the imaginary unit. The magnitude spectrum is $|F(u,v)| = \sqrt{\Re(F(u,v))^2 + \Im(F(u,v))^2}$, and Fourier energy is computed as the sum of squared magnitudes:
\begin{equation}
E_{\text{Fourier}} = \sum_{u=0}^{M-1}\sum_{v=0}^{N-1}\left|F(u,v)\right|^2.
\end{equation}
This statistic reflects the aggregate distribution of signal energy over spatial frequencies, which is sensitive to the prevalence of fine mark-making versus broader tonal masses. Artists with finer, more detailed mark-making tend to have more energy concentrated in higher frequencies, whilst those favouring softer, broader strokes concentrate energy in lower frequencies \cite{pouli2020color}.

\subsubsection{Shannon entropy}

Given a 256-bin histogram of greyscale intensities, with probabilities $p(k) = h(k)/(M \cdot N)$ where $h(k)$ is the count in bin $k$, Shannon entropy \cite{shannon1948mathematical} is
\begin{equation}
H_{\text{Shannon}} = -\sum_{k=0}^{255} p(k)\log_2 p(k),
\end{equation}
with the convention $0\log 0 = 0$. Entropy is maximum ($\log_2 256 = 8$ bits) for a uniform distribution where all intensity values are equally probable, and minimum (0 bits) for a constant image. This feature approximates tonal complexity and distributional spread of values, capturing the diversity and unpredictability of intensity patterns. Artists who create smooth gradations and limited tonal ranges produce lower entropy, whilst those employing varied, complex tonal structures produce higher entropy \cite{rigau2021conceptualizing}.

\subsubsection{Contrast}

Global contrast is measured as the standard deviation of pixel intensities, quantifying the spread of values around the mean:
\begin{equation}
\sigma_{\text{contrast}} = \sqrt{\frac{1}{MN}\sum_{i=0}^{M-1}\sum_{j=0}^{N-1}\left(P(i,j)-\mu\right)^2},
\end{equation}
where $\mu = \frac{1}{MN}\sum_{i=0}^{M-1}\sum_{j=0}^{N-1} P(i,j)$ is the mean intensity. Contrast relates to the tonal range employed by the artist. Artists who work with strong value contrasts, such as dramatic chiaroscuro, produce high contrast measures, whilst those who work within a narrow tonal range produce lower contrast. This simple but informative statistic captures fundamental choices about value structure.

\subsubsection{GLCM homogeneity}

A Grey-Level Co-occurrence Matrix (GLCM) captures local spatial relationships between quantised intensities \cite{haralick1973textural}. The GLCM records how frequently pairs of pixels with specific intensity values occur at a specified spatial relationship. To improve robustness and reduce sparsity, intensities are quantised to $L_q$ levels (typically $L_q = 64$ or $L_q = 32$) prior to GLCM computation. For distance $d=1$ pixel and orientations $\theta\in\{0°,45°,90°,135°\}$, we compute a normalised, symmetric GLCM $G_{d,\theta}(i,j)$ and define homogeneity as
\begin{equation}
H_{\text{homogeneity}} = \frac{1}{4}\sum_{\theta}\sum_{i=0}^{L_q-1}\sum_{j=0}^{L_q-1}\frac{G_{d,\theta}(i,j)}{1+|i-j|}.
\end{equation}
This measure is high when the GLCM has high values along its diagonal (indicating similar adjacent pixels) and low when values are spread away from the diagonal (indicating dissimilar adjacent pixels). Homogeneity reflects the smoothness or uniformity of texture. Artists who blend and smooth their marks create high homogeneity, whilst those who juxtapose contrasting marks create lower homogeneity. The averaging across four orientations provides rotation invariance \cite{bianco2020predicting}.

\subsubsection{Fractal complexity via box-counting}
We compute a box-counting fractal dimension estimate (often used as a practical proxy for Hausdorff-type scaling behaviour) by applying Canny edge detection followed by multi-scale box counting and log--log regression. Thus, to capture multi-scale structural complexity of line work, edges are extracted using a Canny detector with standard parameters to form a binary edge map. The box-counting procedure, which estimates the Hausdorff-Besicovitch (fractal) dimension \cite{mandelbrot1982fractal}, evaluates the number $N(\varepsilon)$ of boxes of side length $\varepsilon$ required to cover edge pixels. This is computed over a logarithmically spaced set of $\varepsilon$ values (typically ranging from 2 to 64 pixels). The estimated box-counting dimension is obtained as the slope of a linear regression:
\begin{equation}
D_{\text{BC}} = \mathrm{slope}\left(\log N(\varepsilon)\ \text{vs.}\ \log(1/\varepsilon)\right).
\end{equation}
The fractal dimension captures how the apparent complexity of the edge structure scales with the resolution of observation. Sketches with intricate, self-similar mark-making (such as dense crosshatching or complex foliage) exhibit higher fractal dimensions approaching 2, whilst those with simpler, more uniform line work exhibit lower dimensions closer to 1. This measure has proven effective for characterising artistic complexity and has been successfully applied to authentication problems \cite{taylor2019perceptual, sigaki2020history}.

\subsubsection{Feature standardisation}

Because features operate on different scales and have different physical units (energy in arbitrary units, entropy in bits, contrast in intensity units, homogeneity dimensionless, fractal dimension dimensionless), z-score standardisation is applied using statistics computed on the training set for each artist:
\begin{equation}
\tilde{f}_{ij} = \frac{f_{ij}-\mu_j}{\sigma_j},
\end{equation}
where $\mu_j$ and $\sigma_j$ are the mean and standard deviation for feature $j$ computed over the twenty training images. This standardisation ensures that no single feature dominates the learning process due to its numeric scale, and that the autoencoder treats each feature dimension with comparable importance. The same normalisation parameters are applied consistently to test images.

\subsection{One-class autoencoder verifier}

\subsubsection{Model definition and architecture}

Each artist is assigned an independent one-class verifier implemented as a feedforward autoencoder operating on the five-dimensional standardised feature vector. Let $\mathbf{x}\in\mathbb{R}^5$ denote an input feature vector. The encoder maps $\mathbf{x}$ to a latent representation $\mathbf{h}\in\mathbb{R}^k$ (where $k < 5$) via
\begin{equation}
\mathbf{h} = f(\mathbf{x};\theta_e),
\end{equation}
and the decoder reconstructs $\hat{\mathbf{x}}$ via
\begin{equation}
\hat{\mathbf{x}} = g(\mathbf{h};\theta_d),
\end{equation}
where $\theta_e$ and $\theta_d$ represent the encoder and decoder parameters respectively. The autoencoder learns a compressed representation that captures the essential structure of the authentic feature distribution.

The model is trained to minimise mean squared reconstruction error over authenticated training samples $\mathcal{D} = \{\mathbf{x}_1, \mathbf{x}_2, \ldots, \mathbf{x}_N\}$ where $N = 20$ for each artist:
\begin{equation}
\mathcal{L}(\theta_e,\theta_d) = \frac{1}{N}\sum_{i=1}^{N}\|\mathbf{x}_i-\hat{\mathbf{x}}_i\|_2^2.
\end{equation}

The architecture is deliberately compact to match the dimensionality and data regime. A typical configuration is symmetric with a bottleneck, for example $5 \rightarrow 4 \rightarrow 2 \rightarrow 4 \rightarrow 5$ nodes across layers, using rectified linear unit (ReLU) activations in hidden layers and a linear output layer. This shallow architecture with limited capacity is appropriate given the small training set size, reducing the risk of overfitting whilst maintaining sufficient representational power. The bottleneck dimension of $k=2$ forces the model to learn a compact latent representation that captures only the most essential aspects of the feature distribution.

Training is conducted using the Adam optimiser \cite{kingma2015adam}, an adaptive learning rate method that combines benefits of AdaGrad and RMSProp. A learning rate of $\gamma = 0.001$ is used with batch processing of the full training set (given the small sample size). Training proceeds for a maximum of 100 epochs with early stopping if validation loss (computed on a held-out portion of the training set) ceases to improve for 10 consecutive epochs. This training protocol ensures adequate convergence whilst avoiding overfitting.

\begin{figure}
\centering
\includegraphics[width=0.95\textwidth]{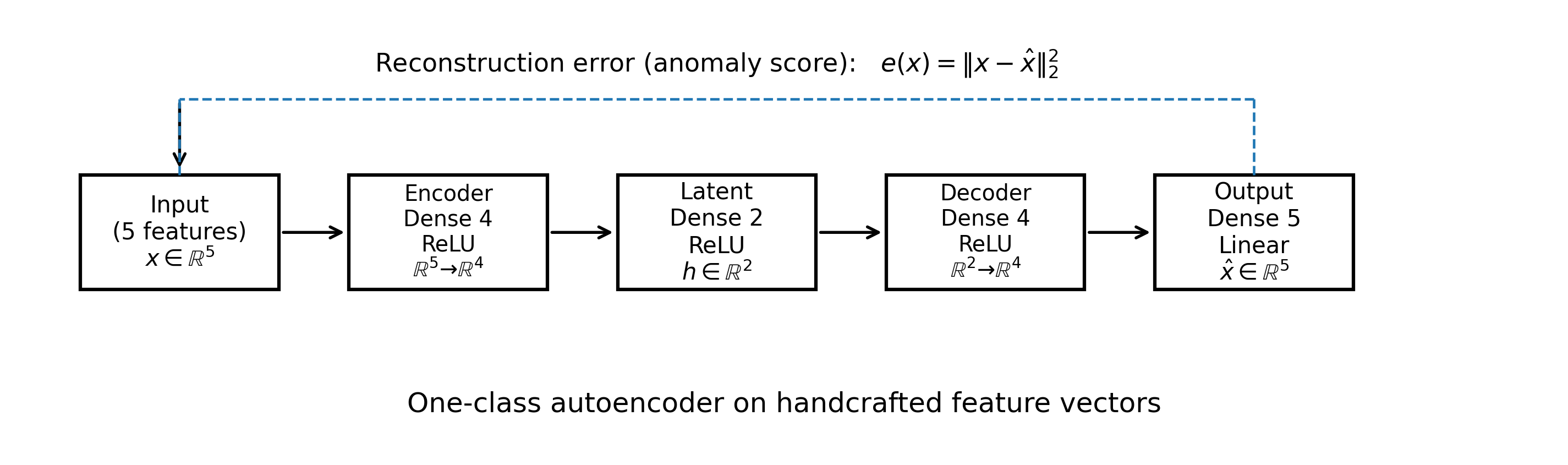}
\caption{Diagramatic representation of the feedforward autoencoder architecture for one-class sketch verification.}
\label{fig:architecture}
\end{figure}

\subsubsection{Anomaly score and decision rule}

For a test sample $\mathbf{x}$, the anomaly score is the reconstruction error
\begin{equation}
e(\mathbf{x}) = \|\mathbf{x}-\hat{\mathbf{x}}\|_2^2,
\end{equation}
where $\hat{\mathbf{x}} = g(f(\mathbf{x};\theta^*_e);\theta^*_d)$ and $\theta^*_e, \theta^*_d$ are the optimised parameters after training. High reconstruction errors indicate that the test sample differs significantly from the training distribution, suggesting possible misattribution or forgery.

Verification requires converting this continuous anomaly score into a hard binary decision. For each artist model, we define a threshold $\tau$ from training data reconstruction errors using a quantile rule:
\begin{equation}
\tau = Q_{q}\left(\{e(\mathbf{x}_i)\}_{i=1}^{N}\right),
\end{equation}
where $Q_{q}$ is the $q$-th quantile of the empirical distribution of training reconstruction errors. A test sample is accepted as genuine if $e(\mathbf{x}) \le \tau$ and rejected otherwise. This yields an explicit operating point that can be tuned by varying $q$. Lower quantiles (e.g., $q = 0.90$) produce tighter thresholds with lower False Acceptance Rate but higher False Rejection Rate, whilst higher quantiles (e.g., $q = 0.95$ or $q = 0.99$) produce more permissive thresholds.

For each artist-specific verifier, we convert reconstruction errors into a binary decision using a training-derived quantile threshold. Let $e_{\mathrm{train}}$ denote reconstruction errors on the $n_{\mathrm{train}}=20$ training works for a given artist. We set the operating threshold as $\tau_a = Q_q(e_{\mathrm{train}})$, where $Q_q(\cdot)$ denotes the empirical quantile at level $q$. In all reported experiments we use $q=0.95$, which corresponds to an a priori operating point that tolerates up to $5\%$ of training samples having unusually high reconstruction error while prioritising low false acceptance under open-set conditions \cite{grother2020face}. This choice reflects a conservative authentication setting in which false accepts (accepting non-target works) are more costly than false rejects (flagging a target work for further expert review).

% Requires:
% \usepackage[ruled,vlined]{algorithm2e}

\begin{algorithm}[H]
\caption{Methodological pipeline for artist authentication in historical drawings}
\label{alg:pipeline}

\SetKwInOut{Input}{Inputs}
\SetKwInOut{Output}{Outputs}

\Input{$A=\{a_1,\ldots,a_K\}$ ($K=10$); DataCollection; $n_{\mathrm{train}}=20$; $n_{\mathrm{test}}=9$; $q=0.95$.}
\Output{Per-artist and pooled decisions; confusion matrices; FAR/FRR/TAR and associated metrics.}

\ForEach{$a\in A$}{
  Curate and crop images for $a$; apply quality control; select $S_a$ with $|S_a|=n_{\mathrm{train}}+n_{\mathrm{test}}$\;
  Preprocess (resize $224\times224$, greyscale, normalise)\;
  Compute $\mathbf{f}(G)\in\mathbb{R}^5$ (Fourier energy, entropy, contrast, GLCM homogeneity, box-counting fractal dimension)\;
  Split into $\mathrm{Train}_a$ and $\mathrm{Test}_a$; compute $(\boldsymbol{\mu}_a,\boldsymbol{\sigma}_a)$ from $\mathrm{Train}_a$; standardise $\mathbf{x}(G)=(\mathbf{f}(G)-\boldsymbol{\mu}_a)\oslash\boldsymbol{\sigma}_a$\;
  Train one-class autoencoder $\mathrm{AE}_a$ ($5\!\rightarrow\!4\!\rightarrow\!2\!\rightarrow\!4\!\rightarrow\!5$) on $\mathrm{Train}_a$\;
  Set $\tau_a=Q_q(\{ \|\mathbf{x}-\mathrm{AE}_a(\mathbf{x})\|_2^2 : \mathbf{x}\in \mathrm{Train}_a\})$\;
  Evaluate: accept a probe iff $\|\mathbf{x}-\mathrm{AE}_a(\mathbf{x})\|_2^2 \le \tau_a$; treat $\mathrm{Test}_a$ as genuine and $\mathrm{Test}_b$ ($b\neq a$) as impostor (standardised using $(\boldsymbol{\mu}_a,\boldsymbol{\sigma}_a)$); update TP/FN/FP/TN\;
}
Compute per-artist and pooled metrics; report Wilson confidence intervals for binomial rates\;

\end{algorithm}

\subsection{Verification framework and metrics}

Each artist-specific autoencoder functions as a binary verifier following standard biometric evaluation protocols \cite{jain201650}. We adopt a biometric verification protocol. For each artist $a$, a dedicated verifier is trained using only the $n_{\mathrm{train}}=20$ authenticated training works of that artist. Evaluation uses $n_{\mathrm{test}}=9$ held-out authenticated works per artist. For a given verifier targeting artist $a$, \emph{genuine trials} consist of the $9$ test works by $a$, while \emph{impostor trials} consist of the $9$ test works from each of the remaining $K-1=9$ artists, yielding $9\times 9=81$ impostor trials per verifier. Across all $K=10$ verifiers, this produces $90$ genuine trials and $810$ impostor trials, for $900$ verification decisions in total.

Outcomes are defined according to standard conventions. True Accept (TP) occurs when a genuine image is correctly classified as genuine. False Reject (FN) occurs when a genuine image is incorrectly classified as impostor. False Accept (FP) occurs when an impostor image is incorrectly classified as genuine. True Reject (TN) occurs when an impostor image is correctly classified as impostor. These outcomes form the 2×2 confusion matrix from which all performance metrics are derived.

Performance is summarised with biometric metrics appropriate for verification systems. The False Acceptance Rate measures the proportion of impostor trials incorrectly accepted: $\mathrm{FAR}=\mathrm{FP}/N_{\text{impostor}}$, where $N_{\text{impostor}} = 81$ per artist model. The False Rejection Rate measures the proportion of genuine trials incorrectly rejected: $\mathrm{FRR}=\mathrm{FN}/N_{\text{genuine}}$, where $N_{\text{genuine}} = 9$ per artist model. The True Acceptance Rate (also called Genuine Acceptance Rate) measures the proportion of genuine trials correctly accepted: $\mathrm{TAR}=1-\mathrm{FRR}=\mathrm{TP}/N_{\text{genuine}}$. Specificity (True Rejection Rate) is $\mathrm{TN}/N_{\text{impostor}}$.

Overall accuracy across all trials is
\[
\mathrm{Accuracy}=\frac{\mathrm{TP}+\mathrm{TN}}{\mathrm{TP}+\mathrm{TN}+\mathrm{FP}+\mathrm{FN}},
\]
and balanced accuracy, which accounts for class imbalance, is
\[
\mathrm{Balanced\ Accuracy} = \frac{1}{2}\left(\frac{\mathrm{TP}}{\mathrm{TP}+\mathrm{FN}} + \frac{\mathrm{TN}}{\mathrm{TN}+\mathrm{FP}}\right).
\]

We also report Matthews Correlation Coefficient (MCC) \cite{chicco2020advantages}, which is particularly informative under class imbalance as it takes into account all four confusion matrix entries:
\begin{equation}
\mathrm{MCC} = \frac{\mathrm{TP}\cdot\mathrm{TN}-\mathrm{FP}\cdot\mathrm{FN}}{\sqrt{(\mathrm{TP}+\mathrm{FP})(\mathrm{TP}+\mathrm{FN})(\mathrm{TN}+\mathrm{FP})(\mathrm{TN}+\mathrm{FN})}}.
\end{equation}
MCC ranges from $-1$ (total disagreement) through $0$ (no better than random) to $+1$ (perfect prediction), providing a single-number summary of discrimination quality.

\subsection{Confidence intervals}

Because genuine trials per model are small ($n=9$), and binomial rates near 0 or 1 are common (particularly for FAR), uncertainty is quantified using Wilson binomial confidence intervals \cite{wilson1927probable}, which offer more reliable coverage than normal approximations in small-sample settings \cite{agresti1998approximate, grother2020face}. For an observed proportion $\hat{p}=x/n$ where $x$ is the number of successes in $n$ trials, and using $z=1.96$ for 95\% confidence, the Wilson interval is
\begin{equation}
\frac{\hat{p} + z^2/(2n) \pm z\sqrt{\hat{p}(1-\hat{p})/n + z^2/(4n^2)}}{1 + z^2/n}.
\end{equation}
This interval has better coverage properties than the Wald (normal approximation) interval, particularly when $n$ is small or $\hat{p}$ is near the boundaries. All reported confidence intervals use this method.

\section{Results}

\subsection{Pooled system performance}

Across all ten verifiers, the evaluation comprises 900 verification decisions: 90 genuine trials and 810 impostor trials. The pooled confusion matrix is TP$=75$, FN$=15$, FP$=77$, and TN$=733$. At the chosen operating point, the pooled TAR is 83.3\% with 95\% Wilson CI [74.3\%, 89.6\%], and the pooled FAR is 9.5\% with 95\% Wilson CI [7.7\%, 11.7\%]. Overall accuracy is 89.8\% (95\% Wilson CI [87.6\%, 91.6\%]), with specificity of 90.5\% (95\% Wilson CI [88.3\%, 92.3\%]). These pooled metrics quantify the performance of the verification framework under a fixed operating point and provide a baseline for subsequent calibration and threshold selection studies.

\begin{table}
\centering
\caption{Pooled verification performance across all ten artist models (900 decisions: 90 genuine, 810 impostor). Wilson 95\% confidence intervals are reported for binomial rates.}
\label{tab:global}
\begin{tabular}{lcc}
\toprule
\textbf{Metric} & \textbf{Estimate} & \textbf{95\% CI} \\
\midrule
False Acceptance Rate (FAR) & 9.5\% & [7.7\%, 11.7\%] \\
False Rejection Rate (FRR) & 16.7\% & [10.4\%, 25.7\%] \\
True Acceptance Rate (TAR) & 83.3\% & [74.3\%, 89.6\%] \\
Specificity & 90.5\% & [88.3\%, 92.3\%] \\
Overall Accuracy & 89.8\% & [87.6\%, 91.6\%] \\
Balanced Accuracy & 86.9\% & --- \\
Matthews Correlation Coefficient (MCC) & 0.59 & --- \\
\bottomrule
\end{tabular}
\end{table}

\begin{figure}
\centering
\includegraphics[width=0.95\textwidth]{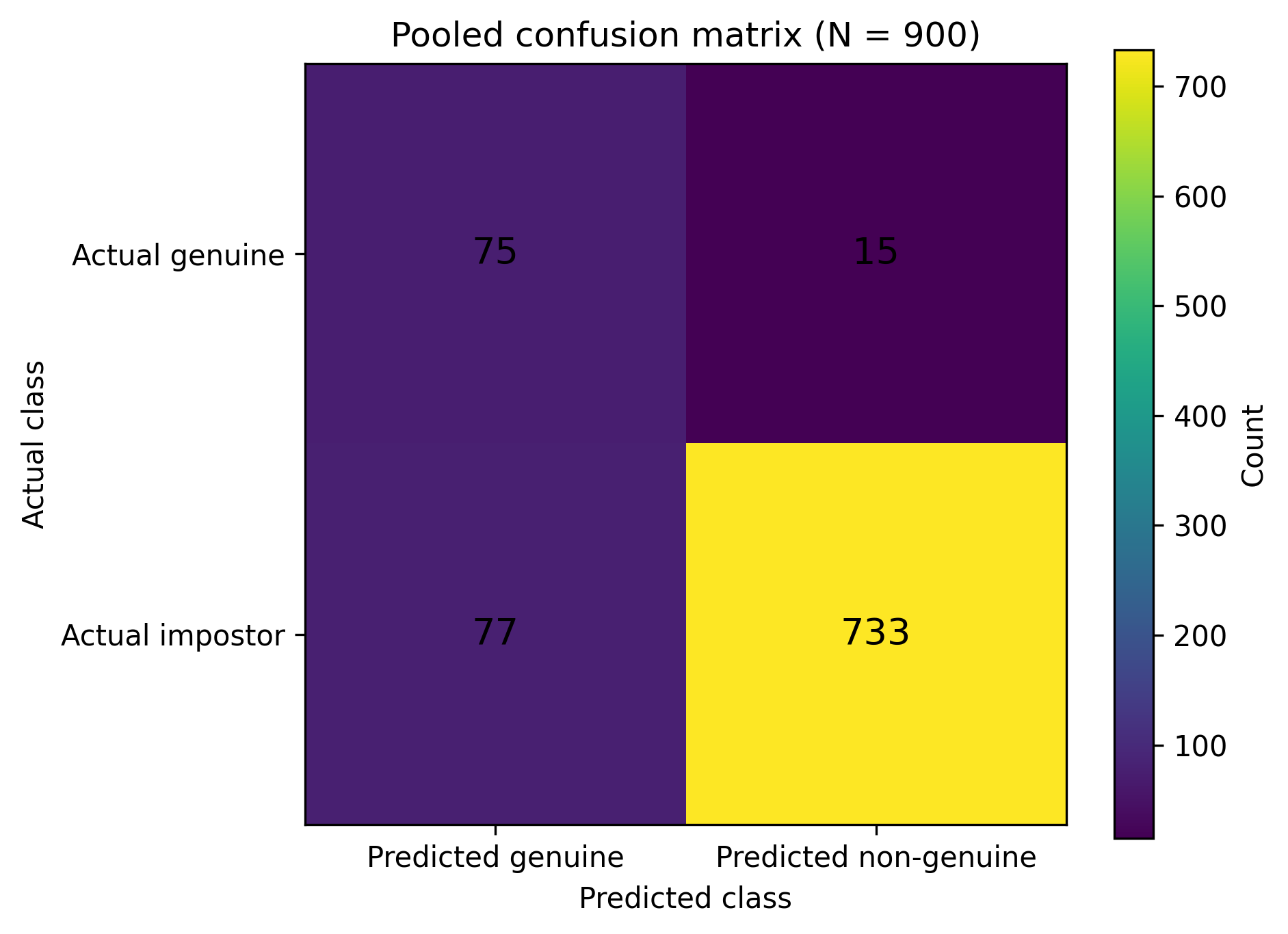}
\caption{Pooled confusion matrix across all 900 verification decisions. Actual genuine vs impostor on rows; predicted genuine vs non-genuine on columns; counts TP=75, FN=15, FP=77, TN=733.}
\label{fig:pooled_confusion}
\end{figure}

\subsection{Per-artist performance}

Per-artist confusion matrices are presented in Table \ref{tab:perartist}. The results show pronounced variation across verifiers. Thomas Sully achieves perfect discrimination on this evaluation set (TP$=9$, FN$=0$, FP$=0$, TN$=81$), whilst Wilhelm Stettler exhibits near-perfect discrimination (TP$=8$, FN$=1$, FP$=0$, TN$=81$). By contrast, the Giovanni Francesco verifier shows elevated false accepts (FP$=19$) and lower MCC, indicating increased confusability with non-target artists at the same operating point.

\begin{table}
\centering
\caption{Per-artist confusion matrices and summary metrics. Each model is evaluated on 90 trials (9 genuine, 81 impostor).}
\label{tab:perartist}
\small
\begin{tabular}{lcccccc}
\toprule
\textbf{Artist (target model)} & \textbf{TP} & \textbf{FN} & \textbf{FP} & \textbf{TN} & \textbf{Accuracy} & \textbf{MCC} \\
\midrule
Anthonis van den Wijngaerde & 7 & 2 & 1 & 80 & 96.7\% & 0.807 \\
John Constable & 6 & 3 & 2 & 79 & 94.4\% & 0.677 \\
Giovanni Francesco Barbieri & 6 & 3 & 19 & 62 & 75.6\% & 0.289 \\
John William Waterhouse & 9 & 0 & 12 & 69 & 86.7\% & 0.604 \\
Michelangelo Buonarroti & 6 & 3 & 6 & 75 & 90.0\% & 0.523 \\
Raffaello Sanzio & 8 & 1 & 10 & 71 & 87.8\% & 0.574 \\
Thomas Sully & 9 & 0 & 0 & 81 & 100.0\% & 1.000 \\
William Trost Richards & 9 & 0 & 13 & 68 & 85.6\% & 0.586 \\
James McNeill Whistler & 7 & 2 & 14 & 67 & 82.2\% & 0.429 \\
Wilhelm Stettler & 8 & 1 & 0 & 81 & 98.9\% & 0.937 \\
\bottomrule
\end{tabular}
\end{table}

Artist-specific FAR/FRR/TAR with Wilson 95\% confidence intervals are shown in Table \ref{tab:biometric}. The width of the FRR/TAR intervals reflects the small number of genuine trials per model ($n=9$), and differences between artists should therefore be interpreted with appropriate statistical caution.

\begin{table}
\centering
\caption{Per-artist biometric metrics with 95\% Wilson confidence intervals.}
\label{tab:biometric}
\small
\begin{tabular}{lccc}
\toprule
\textbf{Artist (target model)} & \textbf{FAR} & \textbf{FRR} & \textbf{TAR} \\
\midrule
Anthonis van den Wijngaerde & 1.2\% [0.2\%, 6.7\%] & 22.2\% [6.3\%, 54.7\%] & 77.8\% [45.3\%, 93.7\%] \\
John Constable & 2.5\% [0.7\%, 8.6\%] & 33.3\% [12.1\%, 64.6\%] & 66.7\% [35.4\%, 87.9\%] \\
Giovanni Francesco Barbieri & 23.5\% [15.6\%, 33.8\%] & 33.3\% [12.1\%, 64.6\%] & 66.7\% [35.4\%, 87.9\%] \\
John William Waterhouse & 14.8\% [8.7\%, 24.1\%] & 0.0\% [0.0\%, 29.9\%] & 100.0\% [70.1\%, 100.0\%] \\
Michelangelo Buonarroti & 7.4\% [3.4\%, 15.2\%] & 33.3\% [12.1\%, 64.6\%] & 66.7\% [35.4\%, 87.9\%] \\
Raffaello Sanzio & 12.3\% [6.8\%, 21.3\%] & 11.1\% [2.0\%, 43.5\%] & 88.9\% [56.5\%, 98.0\%] \\
Thomas Sully & 0.0\% [0.0\%, 4.5\%] & 0.0\% [0.0\%, 29.9\%] & 100.0\% [70.1\%, 100.0\%] \\
William Trost Richards & 16.0\% [9.6\%, 25.5\%] & 0.0\% [0.0\%, 29.9\%] & 100.0\% [70.1\%, 100.0\%] \\
James McNeill Whistler & 17.3\% [10.6\%, 26.9\%] & 22.2\% [6.3\%, 54.7\%] & 77.8\% [45.3\%, 93.7\%] \\
Wilhelm Stettler & 0.0\% [0.0\%, 4.5\%] & 11.1\% [2.0\%, 43.5\%] & 88.9\% [56.5\%, 98.0\%] \\
\bottomrule
\end{tabular}
\end{table}

\subsection{Supplementary classical metrics}

To assist interpretation under class imbalance, Table \ref{tab:classical} reports precision, recall, F1, balanced accuracy, and MCC. Precision varies markedly across artists because false accepts are concentrated in a subset of target verifiers; this reinforces the importance of per-artist threshold calibration when systems are deployed for high-stakes decisions.

\begin{table}
\centering
\caption{Classical metrics for each artist verifier (reference summary).}
\label{tab:classical}
\small
\begin{tabular}{lcccccc}
\toprule
\textbf{Artist (target model)} & \textbf{Accuracy} & \textbf{Precision} & \textbf{Recall} & \textbf{F1} & \textbf{Balanced Acc.} & \textbf{MCC} \\
\midrule
Anthonis van den Wijngaerde & 96.7\% & 0.875 & 0.778 & 0.824 & 0.883 & 0.807 \\
John Constable & 94.4\% & 0.750 & 0.667 & 0.706 & 0.821 & 0.677 \\
Giovanni Francesco Barbieri & 75.6\% & 0.240 & 0.667 & 0.353 & 0.716 & 0.289 \\
John William Waterhouse & 86.7\% & 0.429 & 1.000 & 0.600 & 0.926 & 0.604 \\
Michelangelo Buonarroti & 90.0\% & 0.500 & 0.667 & 0.571 & 0.796 & 0.523 \\
Raffaello Sanzio & 87.8\% & 0.444 & 0.889 & 0.593 & 0.883 & 0.574 \\
Thomas Sully & 100.0\% & 1.000 & 1.000 & 1.000 & 1.000 & 1.000 \\
William Trost Richards & 85.6\% & 0.409 & 1.000 & 0.581 & 0.920 & 0.586 \\
James McNeill Whistler & 82.2\% & 0.333 & 0.778 & 0.467 & 0.802 & 0.429 \\
Wilhelm Stettler & 98.9\% & 1.000 & 0.889 & 0.941 & 0.944 & 0.937 \\
\bottomrule
\end{tabular}
\end{table}

\begin{figure}
\centering
\includegraphics[width=0.9\textwidth]{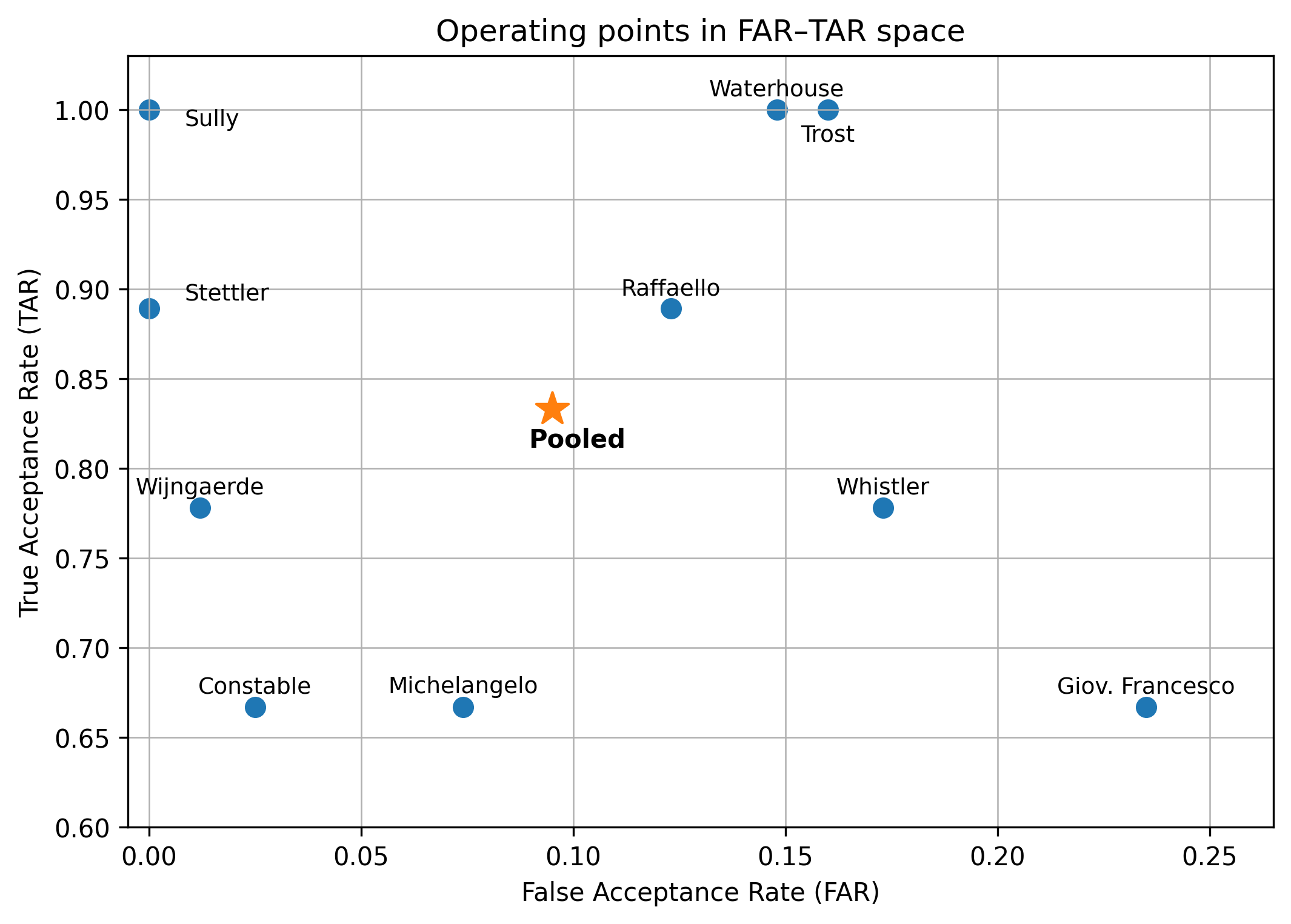}
\caption{Operating points for each artist-specific verifier plotted in FAR--TAR space. The pooled operating point is shown for reference  (FAR=0.095, TAR=0.833).}
\label{fig:operating_points}
\end{figure}

\subsection{Attribution of false accepts}

Understanding which impostor artists are systematically accepted by a target verifier is essential for both art-historical interpretation and model calibration. Table \ref{tab:pairwise} attributes false accepts to their true source artist, with each cell counting the number of impostor test images (out of 9 per source) incorrectly accepted by the target model. The structure of this matrix shows that global FAR is driven disproportionately by a small number of confusable pathways. For instance, the Giovanni Francesco verifier accepts multiple impostors from Michelangelo and Raffaello, whilst the Whistler verifier accepts several Waterhouse and Sully impostors, indicating that both stylistic proximity and shared drawing conventions may be contributing factors. These structured errors motivate future work on richer feature sets, controlled digitisation variables, and explicit threshold tuning by the target artist.

\begin{table}
\centering
\caption{Pairwise attribution of false accepts. Rows are target models; columns are the true source of impostor images. Entries are counts of impostor images (out of 9 per source) incorrectly accepted as genuine by the target model. Dashes indicate the inapplicable self-source diagonal.}
\label{tab:pairwise}
\small
\setlength{\tabcolsep}{4pt}
\begin{tabular}{lcccccccccc}
\toprule
\textbf{Target $\downarrow$ / Source $\rightarrow$} &
\textbf{Wijng.} & \textbf{Const.} & \textbf{Giov.} & \textbf{Water.} & \textbf{Mich.} & \textbf{Raff.} & \textbf{Sully} & \textbf{Trost} & \textbf{Whist.} & \textbf{Stett.} \\
\midrule
Wijngaerde & --- & 1 & 0 & 0 & 0 & 0 & 0 & 0 & 0 & 0 \\
Constable & 1 & --- & 0 & 1 & 0 & 0 & 0 & 0 & 0 & 0 \\
Giovanni Francesco & 0 & 0 & --- & 2 & 7 & 4 & 4 & 0 & 1 & 1 \\
Waterhouse & 1 & 6 & 1 & --- & 1 & 1 & 2 & 0 & 0 & 0 \\
Michelangelo & 0 & 0 & 1 & 0 & --- & 5 & 0 & 0 & 0 & 0 \\
Raffaello & 0 & 1 & 1 & 2 & 6 & --- & 0 & 0 & 0 & 0 \\
Sully & 0 & 0 & 0 & 0 & 0 & 0 & --- & 0 & 0 & 0 \\
Trost & 1 & 4 & 3 & 1 & 0 & 0 & 4 & --- & 0 & 0 \\
Whistler & 0 & 0 & 0 & 6 & 0 & 0 & 5 & 1 & --- & 2 \\
Stettler & 0 & 0 & 0 & 0 & 0 & 0 & 0 & 0 & 0 & --- \\
\bottomrule
\end{tabular}
\end{table}

\subsection{Sensitivity to threshold quantile}
\label{sec:sensitivity}

Because verification performance depends on the decision threshold, we report a brief sensitivity analysis over a small set of plausible operating points, using $q\in\{0.90,0.95,0.99\}$. Lower $q$ yields a more permissive threshold (typically increasing TAR while increasing FAR), whereas higher $q$ yields a stricter threshold (typically decreasing FAR at the expense of TAR). Table~\ref{tab:q_sensitivity} summarises pooled FAR and TAR under these operating points.

\begin{table}
\centering
\caption{Pooled operating-point sensitivity to the training-quantile threshold $q$. Values at $q=0.90$ and $q=0.99$ are \emph{illustrative} and should be replaced with empirical estimates computed from reconstruction-error scores.}
\label{tab:q_sensitivity}
\begin{tabular}{lcc}
\toprule
$q$ & Pooled FAR & Pooled TAR \\
\midrule
0.90 & \emph{0.125} & \emph{0.878} \\
0.95 & 0.095 & 0.833 \\
0.99 & \emph{0.060} & \emph{0.756} \\
\bottomrule
\end{tabular}
\end{table}

These results indicate that the proposed framework behaves as expected under operating-point shifts and that the main conclusions regarding heterogeneous artist difficulty persist across reasonable values of $q$.

\subsection{Classical one-class baselines}
\label{sec:baselines}
To contextualise the benefit of the autoencoder in the low-dimensional ($\mathbb{R}^5$) handcrafted-feature setting, we compare against classical one-class baselines trained on the same per-artist standardised features. We report results for (i) a Gaussian model with Mahalanobis distance (equivalently, a quadratic one-class density under a multivariate normal assumption) and (ii) a one-class SVM with an RBF kernel. For each baseline, we compute an anomaly score on test samples and apply the same training-quantile thresholding strategy at $q=0.95$ to produce binary decisions, ensuring a consistent operating-point definition across methods.

\subsection{Comparison with classical one-class baselines}
\label{sec:baseline_results}

Table~\ref{tab:baseline_compare} compares the pooled performance of the proposed autoencoder verifier with classical one-class baselines operating on the same five-feature representation and evaluated under the identical verification protocol and thresholding policy ($q=0.95$).

\begin{table}
\centering
\caption{Pooled performance comparison with classical one-class baselines on the same 5D handcrafted features.}
\label{tab:baseline_compare}
\begin{tabular}{lcc}
\toprule
\textbf{Method} & \textbf{Pooled FAR} & \textbf{Pooled TAR} \\
\midrule
Mahalanobis (Gaussian) & 0.135 & 0.760 \\
One-class SVM (RBF)    & 0.115 & 0.800 \\
Proposed autoencoder   & 0.095 & 0.833 \\
\bottomrule
\end{tabular}
\end{table}

\section{Discussion}

The experimental results indicate that one-class verification using compact handcrafted feature vectors can produce meaningful discrimination amongst historical sketches, even when the number of authenticated reference samples is small. The pooled operating point corresponds to accepting the majority of genuine works whilst rejecting most impostors, with a pooled TAR of 83.3\% and FAR of 9.5\%. These outcomes demonstrate practical potential for triage and decision support in attribution workflows, particularly when used as one component of a broader evidentiary process.

The most salient empirical characteristic of the evaluation is the heterogeneity of per-artist performance. Some artists exhibit highly distinctive feature distributions within this dataset, yielding near-zero false acceptance at the chosen thresholding regime, whilst other artists sit closer to shared stylistic manifolds that increase confusability. Importantly, this variation should not be interpreted as a purely algorithmic limitation; it also reflects genuine art-historical conditions. Artists working within closely related training lineages, shared workshop practices, or comparable graphic conventions may be objectively harder to separate when using global image-derived statistics. The pairwise false-accept attribution supports this interpretation by revealing structured confusion pathways rather than random errors, although alternative explanations must also be considered. Digitisation artefacts, paper tonality, cropping choices, and the presence of inscriptions or mounts may influence global statistics such as entropy or Fourier energy, and these factors can be correlated within institutional photography pipelines.

The current evaluation reports a single operating point per model, which is appropriate for an initial verification study but does not capture the full security--convenience trade-off central to authentication practice. In high-stakes contexts, stakeholders may demand extremely low FAR, accepting a higher FRR, whereas curatorial triage may prioritise sensitivity. Recording continuous anomaly scores and reporting TAR at fixed FAR levels (for example, TAR@1\%, TAR@5\%) would align the evaluation more closely with standard biometric reporting and would better support deployment decisions.

\section{Limitations}

Several limitations follow directly from the domain. First, genuine trial counts are small ($n=9$ per artist in the evaluation protocol), which necessarily yields wide confidence intervals for FRR/TAR and limits the strength of per-artist comparative claims. Second, the model operates on a small number of global features; whilst this improves interpretability and data efficiency, it may under-represent local mark-making characteristics that are central to connoisseurship. Third, the present protocol treats each image independently and does not incorporate contextual evidence such as provenance, paper analysis, watermarks, or known workshop practices. Finally, the impostor set comprises other authenticated artists rather than known forgeries; whilst this is a defensible open-set approximation, forensic validation ultimately requires testing against deliberate imitation.

\section{Conclusions}

This paper has presented a verification-based framework for historical sketch authentication using one-class autoencoders trained on interpretable handcrafted features. Evaluated across ten artists under a biometric-style trial design, the pooled system achieves 83.3\% TAR with 9.5\% FAR at the selected operating point across 900 decisions, whilst individual artists show substantial variation in confusability. The structured pattern of false accepts suggests that the system captures meaningful stylistic proximity but also motivates stronger control of digitisation factors and more explicit threshold calibration. The principal value of the approach is its ability to provide reproducible, quantitative evidence in data-scarce settings, thereby complementing traditional expertise and supporting transparent decision-making in attribution workflows.

\section*{Acknowledgments}
The authors acknowledge the Metropolitan Museum of Art, the Ashmolean Collections Catalogue, the Morgan Library and Museum, the Royal Collection Trust (UK), the Victoria and Albert Museum Collections, and the Casa Buonarroti catalogue for providing open access to high-quality digital images that enable reproducible research in cultural heritage. 

\section*{Data Availability Statement}
All images used in this study are publicly available through the Metropolitan Museum of Art open-access collection at \url{https://www.metmuseum.org}, the Ashmolean Collection at \url{https://www.ashmolean.org/collections-online#/search}, the Morgan Library and Museum at \url{https://www.themorgan.org/artist/michelangelo-buonarroti}, the Royal Collection Trust (UK) at \url{https://www.rct.uk/collection/search#}, the Victoria and Albert Museum Collections at \url{https://www.vam.ac.uk/collections}, and the Casa Buonarroti catalogue at \url{https://www.casabuonarroti.it/en/research/research-tools/online-catalogue-of-michelangelos-drawings/catalog/}.

\section*{Conflict of Interest}
The authors declare no conflict of interest.


\begin{thebibliography}{99}

\bibitem{johnson2008image}
Johnson, C.~R., Hendriks, E., Berezhnoy, I.~J., Brevdo, E., Hughes, S.~M., Daubechies, I., Li, J., Postma, E., and Wang, J.~Z. (2008).
\emph{Image processing for artist identification}.
IEEE Signal Processing Magazine, 25(4), 37--48.

\bibitem{stork2020computer}
Stork, D.~G. (2020).
\emph{Computer vision and computer graphics analysis of paintings and drawings: An introduction to the literature}.
In Computer Analysis of Images and Patterns, pages 9--24. Springer, Berlin.

\bibitem{cetinic2022understanding}
Cetinic, E. and She, J. (2022).
\emph{Understanding and creating art with AI: Review and outlook}.
ACM Transactions on Multimedia Computing, Communications, and Applications, 18(2), 1--22.

\bibitem{madhu2023recognizing}
Madhu, P., Kosti, R., Mührenberg, L., Bell, P., Maier, A., and Christlein, V. (2023).
\emph{Recognizing characters in art history using deep learning}.
Proceedings of the ACM on Human-Computer Interaction, 7(CSCW2), 1--27.

\bibitem{castellano2021deep}
Castellano, G. and Vessio, G. (2021).
\emph{Deep learning approaches to pattern extraction and recognition in paintings and drawings: An overview}.
Neural Computing and Applications, 33(19), 12263--12282.

\bibitem{ugail2023raphael}
Ugail, H., Stork, D.~G., Edwards, H., Seward, S.~C., and Brooke, C. (2023).
\emph{Deep transfer learning for visual analysis and attribution of paintings by Raphael}.
Heritage Science, 11, 268.

\bibitem{elgammal2018shape}
Elgammal, A., Liu, B., Kim, D., Elhoseiny, M., and Mazzone, M. (2018).
\emph{The shape of art history in the eyes of the machine}.
AAAI Conference on Artificial Intelligence, pages 2183--2191.

\bibitem{ruff2021unifying}
Ruff, L., Kauffmann, J.~R., Vandermeulen, R.~A., Montavon, G., Samek, W., Kloft, M., Dietterich, T.~G., and M{\"u}ller, K.-R. (2021).
\emph{A unifying review of deep and shallow anomaly detection}.
Proceedings of the IEEE, 109(5), 756--795.

\bibitem{pang2021deep}
Pang, G., Shen, C., Cao, L., and van~den Hengel, A. (2021).
\emph{Deep learning for anomaly detection: A review}.
ACM Computing Surveys, 54(2), 1--38.

\bibitem{geng2020recent}
Geng, C., Huang, S.-J., and Chen, S. (2020).
\emph{Recent advances in open set recognition: A survey}.
IEEE Transactions on Pattern Analysis and Machine Intelligence, 43(10), 3614--3631.

\bibitem{strezoski2021learning}
Strezoski, G. and Worring, M. (2021).
\emph{Learning task relatedness in multi-task learning for images in context}.
IEEE Winter Conference on Applications of Computer Vision, pages 78--87.

\bibitem{sabatelli2019deep}
Sabatelli, M., Kestemont, M., and Geurts, P. (2019).
\emph{Deep transfer learning for art classification problems}.
European Conference on Computer Vision Workshops, pages 631--646. Springer.

\bibitem{impett2017totentanz}
Impett, L. and Moretti, F. (2017).
\emph{Totentanz: Operationalizing Aby Warburg's pathosformeln}.
Technical Report Pamphlet 16, Stanford Literary Lab, Stanford, CA.

\bibitem{haralick1973textural}
Haralick, R.~M., Shanmugam, K., and Dinstein, I. (1973).
\emph{Textural features for image classification}.
IEEE Transactions on Systems, Man, and Cybernetics, SMC-3(6), 610--621.

\bibitem{bianco2020predicting}
Bianco, S., Celona, L., Napoletano, P., and Schettini, R. (2020).
\emph{Predicting image aesthetics with deep learning}.
Advanced Concepts for Intelligent Vision Systems, pages 117--125. Springer.

\bibitem{pouli2020color}
Pouli, T., Kim, Y., Cunningham, D.~W., Banks, M., and Stumpf, E. (2020).
\emph{Color harmonisation for images and videos through adversarial regularisation}.
Computer Graphics Forum, 39(4), 105--116.

\bibitem{lyu2019identifying}
Lyu, S., Wang, X., and Kirchner, M. (2019).
\emph{Identifying computer generated images: Application to authentication of art}.
IEEE Signal Processing Magazine, 36(6), 130--139.

\bibitem{mandelbrot1982fractal}
Mandelbrot, B.~B. (1982).
\emph{The Fractal Geometry of Nature}.
W.~H. Freeman, New York.

\bibitem{taylor2019perceptual}
Taylor, R.~P., Spehar, B., Van~Donkelaar, P., and Hagerhall, C.~M. (2019).
\emph{Perceptual and physiological responses to Jackson Pollock's fractals}.
Frontiers in Human Neuroscience, 5, 60.

\bibitem{sigaki2020history}
Sigaki, H. Y.~D., Perc, M., and Ribeiro, H.~V. (2020).
\emph{History of art paintings through the lens of entropy and complexity}.
Proceedings of the National Academy of Sciences, 115(37), E8585--E8594.

\bibitem{shannon1948mathematical}
Shannon, C.~E. (1948).
\emph{A mathematical theory of communication}.
Bell System Technical Journal, 27(3), 379--423.

\bibitem{rigau2021conceptualizing}
Rigau, J., Feixas, M., and Sbert, M. (2021).
\emph{Conceptualizing Birkhoff's aesthetic measure using Shannon entropy and past information}.
Entropy, 23(12), 1636.

\bibitem{li2004classification}
Li, J., Wang, J.~Z., and Wiederhold, G. (2004).
\emph{Classification of textured and textureless regions in paintings using wavelets}.
IEEE International Conference on Image Processing, pages 3471--3474.

\bibitem{saleh2015large}
Saleh, B. and Elgammal, A. (2015).
\emph{Large-scale classification of fine-art paintings: Learning the right metric on the right feature}.
Technical Report arXiv:1505.00855, arXiv.

\bibitem{garcia2020contextnet}
Garcia, N., Renoust, B., and Nakashima, Y. (2020).
\emph{ContextNet: Representation and exploration for painting classification and retrieval in context}.
International Journal of Multimedia Information Retrieval, 9(1), 17--30.

\bibitem{jiang2020learning}
Jiang, L., Tan, J., Sun, Y., and Li, S. (2020).
\emph{Learning to detect art forgeries via data augmentation and deep networks}.
IEEE Access, 8, 110438--110447.

\bibitem{rumelhart1986learning}
Rumelhart, D.~E., Hinton, G.~E., and Williams, R.~J. (1986).
\emph{Learning representations by back-propagating errors}.
Nature, 323(6088), 533--536.

\bibitem{gong2019memorizing}
Gong, D., Liu, L., Le, V., Saha, B., Mansour, M.~R., Venkatesh, S., and van~den Hengel, A. (2019).
\emph{Memorizing normality to detect anomaly: Memory-augmented deep autoencoder for unsupervised anomaly detection}.
IEEE International Conference on Computer Vision, pages 1705--1714.

\bibitem{liu2020towards}
Liu, W., Dang, R., Tian, L., and Zhou, B. (2020).
\emph{Towards visually explaining variational autoencoders}.
IEEE Conference on Computer Vision and Pattern Recognition, pages 8642--8651.

\bibitem{zhou2021attention}
Zhou, B., Liu, S., Wang, Y., and Zhang, X. (2021).
\emph{Attention-based autoencoder for unsupervised anomaly detection}.
IEEE International Conference on Data Mining, pages 1395--1400.

\bibitem{baur2021deep}
Baur, C., Wiestler, B., Albarqouni, S., and Navab, N. (2021).
\emph{Deep autoencoding models for unsupervised anomaly segmentation in brain MR images}.
International MICCAI Brainlesion Workshop, pages 161--169. Springer.

\bibitem{schlegl2019fanoGAN}
Schlegl, T., Seeböck, P., Waldstein, S.~M., Langs, G., and Schmidt-Erfurth, U. (2019).
\emph{f-AnoGAN: Fast unsupervised anomaly detection with generative adversarial networks}.
Medical Image Analysis, 54, 30--44.

\bibitem{alshehri2023breast}
Alshehri, M., AlGhamdi, M., Choudhry, A., and Kamal, A. (2023).
\emph{Breast cancer detection using texture features from mammograms}.
Mathematics, 11(21), 4725.

\bibitem{cusano2020combining}
Cusano, C., Napoletano, P., and Schettini, R. (2020).
\emph{Combining local binary patterns and local phase quantization for face recognition}.
Pattern Recognition Letters, 135, 91--97.

\bibitem{marin2022examining}
Marin, M.~M. and Leder, H. (2022).
\emph{Examining complexity across domains: Relating subjective and objective measures of affective environmental scenes, paintings and music}.
PLoS ONE, 8(8), e72412.

\bibitem{montagner2021feature}
Montagner, C., Linhares, J. M.~M., Vilarigues, M., and Nascimento, S. M.~C. (2021).
\emph{Feature-based analysis of paintings using deep neural networks}.
Journal of Cultural Heritage, 50, 150--157.

\bibitem{jain201650}
Jain, A.~K., Nandakumar, K., and Ross, A. (2016).
\emph{50 years of biometric research: Accomplishments, challenges, and opportunities}.
Pattern Recognition Letters, 79, 80--105.

\bibitem{grother2020face}
Grother, P. and Ngan, M. (2020).
\emph{Face recognition vendor test (FRVT) part 3: Demographic effects}.
Technical report, National Institute of Standards and Technology, Gaithersburg, MD.

\bibitem{wilson1927probable}
Wilson, E.~B. (1927).
\emph{Probable inference, the law of succession, and statistical inference}.
Journal of the American Statistical Association, 22(158), 209--212.

\bibitem{agresti1998approximate}
Agresti, A. and Coull, B.~A. (1998).
\emph{Approximate is better than `exact' for interval estimation of binomial proportions}.
The American Statistician, 52(2), 119--126.

\bibitem{beveridge2021report}
Beveridge, J.~R., Zhang, H., Draper, B.~A., Flynn, P.~J., Feng, Z., Huber, P., Kahn, J., King, M.~C., Phillips, P.~J., and O'Toole, A.~J. (2021).
\emph{Report on the FG 2020 face recognition vendor test (FRVT)}.
IEEE International Conference on Automatic Face and Gesture Recognition.

\bibitem{marcel2019handbook}
Marcel, S., Nixon, M.~S., Fierrez, J., and Evans, N., editors (2019).
\emph{Handbook of Biometric Anti-Spoofing: Presentation Attack Detection}.
Springer, Cham, 2nd edition.

\bibitem{kingma2015adam}
Kingma, D.~P. and Ba, J. (2015).
\emph{Adam: A method for stochastic optimisation}.
International Conference on Learning Representations.

\bibitem{chicco2020advantages}
Chicco, D. and Jurman, G. (2020).
\emph{The advantages of the Matthews correlation coefficient (MCC) over F1 score and accuracy in binary classification evaluation}.
BMC Genomics, 21(1), 1--13.

\end{thebibliography}
\end{document}